\pdfoutput=1

\documentclass[11pt]{article}

\usepackage[]{EMNLP2023}

\usepackage{times}
\usepackage{latexsym}

\usepackage[T1]{fontenc}

\usepackage[utf8]{inputenc}

\usepackage{microtype}

\usepackage{inconsolata}

\usepackage{amssymb}
\usepackage{pifont}
\newcommand{\cmark}{\ding{51}}%
\newcommand{\xmark}{\ding{55}}%
\usepackage{multirow}
\usepackage{array}
\usepackage{booktabs}
\usepackage{comment}
\title{Killing Two Flies with One Stone: An Attempt to Break LLMs Using English$\rightarrow$Icelandic Idioms and Proper Names}

 \author{Bjarki Ármannsson, Hinrik Hafsteinsson, Atli Jasonarson, Steinþór Steingrímsson \\
         The Árni Magnússon Institute for Icelandic Studies\\
         Reykjavík, Iceland\\\texttt{bjarki.armannsson,hinrik.hafsteinsson,atli.jasonarson,}\\\texttt{steinthor.steingrimsson@arnastofnun.is}}

\begin{document}
\maketitle
\begin{abstract}
This paper presents the submission of the Árni Magnússon Institute's team to the WMT24 test suite subtask, focusing on idiomatic expressions and proper names for the English$\rightarrow$Icelandic translation direction. 

Intuitively and empirically, idioms and proper names are known to be a significant challenge for modern translation models. We create two different test suites. The first evaluates the competency of MT systems in translating common English idiomatic expressions, as well as testing whether systems can distinguish between those expressions and the same phrases when used in a literal context. The second test suite consists of place names that should be translated into their Icelandic exonyms (and correctly inflected) and pairs of Icelandic names that share a surface form between the male and female variants, so that incorrect translations impact meaning as well as readability. 

The scores reported are relatively low, especially for idiomatic expressions and place names, and indicate considerable room for improvement.
\end{abstract}

\section{Introduction}

Significant advances in machine translation have in recent years been achieved by integrating Large Language Models (LLMs) into neural translation systems \citep{xu2024}. Careful analysis, however, has repeatedly shown that despite recording higher scores and producing text with greater fluency compared to previous state-of-the-art neural systems, the translations produced by LLMs are still far from perfect and can include significant biases, misinformation and hallucinations \cite{hendy2023goodgptmodelsmachine}, half-hidden in the impressive-looking output. Aiming to expose ``weaknesses and serious flaws'' of these systems that might otherwise get ``hidden in the average'', the theme of this year's WMT test suite subtask is ``Help us break LLMs'', with organizers asking for custom test sets focusing on phenomena that can provide specific challenges for LLM-based systems. This paper describes the efforts of the Árni Magnússon Institute's team to pick holes in otherwise seemingly fluent English$\rightarrow$Icelandic translations.

We experiment with two main features we believe should prove particularly challenging for English$\rightarrow$Icelandic LLM-based machine translation systems; idiomatic expressions and proper names. More specifically, we focus on:
\begin{itemize}
    \item \textbf{Idiomatic expressions in English and their literal counterparts:} In the first of our two test sets, we investigate idiomatic expressions in English which do not directly translate to Icelandic. Where possible, we also include `inverse' examples of usage in a literal form (as in ``Are you supposed to \textbf{chew the fat} from steak?'' or ``Blow into the balloon and \textbf{tie the knot} without letting the air out.'') to give an idea of the translation models' ability to correctly switch between literal and non-literal translations of the same phrase, depending on context.
    \item \textbf{Proper names:} In our second test set, we also consider names of both people and places. We carefully curate a list of city and area names in English that should be translated to their common Icelandic names (and correctly inflected). We then include a list of simple sentences containing both Icelandic and English given names. For the Icelandic names, we observe whether they are correctly inflected in the Icelandic text (which impacts not only the text's readability, but also its meaning). Common English names, meanwhile, are included to test that the models don't `translate' them to Icelandic -- i.e. alter them in some unintended way.
\end{itemize}

We release our test suites and evaluation code for others to build on and to allow for further comparison between future models in these categories.\footnote{\url{https://github.com/stofnun-arna-magnussonar/idioms_names_test_suite}}

\section{Related Work}

Idiomatic expressions (and multi-word expressions (MWEs) in general) have been the focus of much work in the field of machine translation in recent years and the construction of impressive idiom datasets has been carried out for many other languages and language pairs. See e.g. \citet{stap-etal-2024-fine} for English$\leftrightarrow$German and Russian$\rightarrow$English, \citet{tang2022petciparallelenglishtranslation} for Chinese$\rightarrow$English, \citet{fadaee-etal-2018-examining} for English$\leftrightarrow$German and \citet{haagsma-etal-2020-magpie} and \citet{adewumi-etal-2022-potential} for monolingual datasets of English idiomatic expressions. 

\citet{macketanz-EtAl:2022:LREC} include idioms among many other interesting linguistic phenomena in their dataset for English$\leftrightarrow$German and English$\rightarrow$Russian and we took some inspiration from their work when deciding on our scoring format. \citet{20.500.12537/275} list Icelandic idioms with English equivalents, this dataset is described and discussed in more detail in \citet{2024_idiamaticexpressions_clarin}. We are not aware of any dataset for the English$\rightarrow$Icelandic translation direction published previous to our work.

In recent years, the emergence of LLMs has led to work investigating how they handle the translations of idioms and MWEs compared with previous models. \citet{raunak-etal-2023-gpts}, using measures of `literalness', find that GPT models produce less literal translations between English and German, Chinese, and Russian than previous neural models, a difference most pronounced in the case of idiomatic expressions. Finally, \citet{shwartz-2021-long} provide an accessible overview of the kinds of problems posed by MWEs for language models in general.

\section{Methodology}
\label{methodology}

\subsection{Idiomatic Expressions}
\label{idmethod}
\begin{table*}
\centering
\begin{tabular}[c]{m{5cm}|m{5cm}|m{3cm}}
\hline
\textbf{Source sentence} &\textbf{Possible translations} & \textbf{Evaluation}\\
\hline
\multirow{2}{*}{Why Fleabag is \textbf{in the pink}!} & Fleabag er \textbf{\textit{í góðum málum}}! & \cmark (Positive match) \\
& Fleabag er \textbf{í \underline{bleiku}}! & \xmark  \; (Negative match)\\
\hline
\multirow{2}{11em}{The young woman \textbf{in the pink} 
continued to throw punches [...]} & Unga konan \textbf{í góðum málum} lét hnefana tala áfram [...] & \xmark  \; (No positive match) \\
 & Unga konan \textbf{í \underline{bleiku} fötunum} lét hnefana tala áfram [...]\ & \cmark (Positive match) \\
\hline
\end{tabular}
\caption{\label{idioms-table} 
Fabricated example translations into Icelandic of two English sentences containing the phrase ``in the pink'', both from our test suite. The first English sentence uses the phrase in an idiomatic sense (meaning \textit{in good health} or \textit{in a state of well-being}) and the second seemingly in a literal sense (the full context, not included in the table, is: ``before another wades in''). For the idiomatic sentence, we automatically mark it as correct if a match is found from a list of possible Icelandic translations (here the phrase ``í góðum málum'') \textit{and} no match is found from a list of negative matches (here the lexeme ``bleikur'', meaning \textit{pink}). For the literal sentence, meanwhile, some form of ``bleikur'' is required for a correct marking.
}
\end{table*}

We make use of the set of potential idiomatic expressions defined in the PIE Corpus \citep{adewumi-etal-2022-potential} and, for each expression we use, extract two examples of usage from the NewsCrawl corpus of WMT 2023 \cite{kocmi-etal-2023-findings}\footnote{\url{https://data.statmt.org/news-crawl/}} For our purposes, we narrow the PIE set down from 591 expressions to 199. Our aim was to remove those we deem too rare or obscure to be truly relevant (e.g. \textit{horses for courses} or \textit{monkey's uncle}) for model comparison, expressions which directly (or more or less directly) translate between English and Icelandic (e.g. ``open the floodgates'' has an Icelandic equivalent, ``opna flóðgáttirnar'') and those for which we find no example usage in the NewsCrawl corpus. We make the number of expressions an even 200 by adding one that was not in the PIE corpus: ``kill two birds with one stone''.

Each of the 400 example sentences - two examples for each of the 200 selected idioms - is then manually reviewed to make sure that the relevant idiomatic expression is being used in the intended, non-literal sense. To further increase the difficulty of the task (though still keeping it trivial for fluent human speakers of Icelandic and English), we also try and test the models on their ability to translate the words in these expressions literally when appropriate. We include 223 additional example sentences, for as many expressions as we were able, where the expression is used in a literal sense (or in a few cases, very slightly altered to try and exploit the likelihood bias of LLMs).\footnote{Early inspiration for this project was provided by the one idiomatic expression we added from outside the PIE corpus: ``kill two birds with one stone''. We noticed a prominent online translation service correctly translated this to the equivalent Icelandic phrase, ``slá tvær flugur í einu höggi'' (lit. \textit{hit two flies in one strike}), whereas a phrase like ``He killed two birds yesterday'' would be wrongly translated as ``Hann drap tvær flugur í gær'' (lit. \textit{He killed two flies yesterday}), exposing a weakness particular to neural and LLM-based systems. Indeed, four of the systems tested here made this particular mistake.} These examples are largely taken from the NewsCrawl corpus but synthetic in some cases. 

To evaluate the models' performance, we construct two `positive' sets of Icelandic word forms or multiword expressions for each idiom. One set contains words that we would expect to find in a literal translation of the phrase, the other words or phrases that could be expected to appear in a suitable, non-literal translation of the idiomatic expression. In many cases, we also construct `negative' sets of words that instantly lead to a sentence being marked incorrect, such as the Icelandic words for ``weather'' or ``pink'' for idiomatic translations of the phrases ``under the weather'' and ``in the pink''. An Icelandic translation of an example sentence in English is marked as correct if it contains any of the words in the set of `positive' words (in any lexical form) \textbf{and} it contains none of the words in the set of `negative' words (see Table \ref{idioms-table}). 

During our manual evaluation, we further whittled down our set as we decided a few sentences we had decided to include were actually not testing what they were meant to test (as some were, for instance, more linguistically acceptable when translated directly into Icelandic than we originally felt during the construction of our test set). We removed a total of 25 sentences this way, bringing the total of `idiomatic' examples in our set to 397 and the total of `literal' examples to 201. Note that although these examples were removed after we received their translations from the tested models, they are not included in our scoring.


\subsection{Proper Names}
\label{propernames}
\begin{table*}
\centering
\begin{tabular}{c | c | c}
\hline
\textbf{Source sentence} & \textbf{Possible translations} & \textbf{Evaluation}\\
\hline
\multirow{2}{*}{\textbf{Helgi} dreams of flying} & \textbf{Helgi} dreymir um að fljúga & \xmark \; (Ungrammatical) \\
& \textbf{Helga} dreymir um að fljúga & \cmark \\
\hline
\multirow{2}{*}{\textbf{Helga} dreams of flying} & \textbf{Helga} dreymir um að fljúga & \xmark  \; (Refers to Helgi, not Helga) \\
 & \textbf{Helgu} dreymir um að fljúga & \cmark \\
\hline
\end{tabular}
\caption{\label{propernames-table} 
Examples of possible translations of the phrase ``dreams of flying''. In Icelandic, the verb ``dreyma'' (\textit{to dream}) takes a subject argument in the accusative case, which requires a translation system to alter the form of the given name in the English text. Left unaltered in Icelandic, the male name ``Helgi'' renders the sentence ungrammatical and the female name ``Helga'' would cause the reader to interpret the sentence to refer to a male called Helgi instead.
}
\end{table*}

For our testing of place names, we construct our own list of 52 names of cities and areas that we argue would be highly unusual not to translate into their Icelandic names.\footnote{There exist context-dependent exceptions to this, of course, such as the name of a sport club or particular institution from a certain city. Our example sentences, however, refer clearly to the cities in general.} 

As a reference when collecting our place names, we make use of Wikipedia's list of Icelandic exonyms.\footnote{\url{https://en.wikipedia.org/wiki/Icelandic_exonyms}} We use only a small subset of that list, however. Aiming to err on the side of caution, we try to include only place names where native speakers would be in more or less complete agreement to apply their Icelandic names rather than the ones used in English (e.g. the name ``Kaupmannahöfn'' for Copenhagen is invariably used, whereas ``Lundúnir'' for London is very rare and mostly used in a colourful or joking manner.\footnote{One anonymous reviewer asked whether we had considered incorporating a native speaker survey in order to validate our choices. While the suggestion is certainly a good one, it is beyond the scope of this particular work.} We also leave out cases where the differences between the names used in English and in Icelandic only have to do with pronunciation or minor differences in spelling. In addition to the Icelandic exonyms we select, we make sure to also include several examples of cities where the local name is the one more generally used and English speakers use a rarer (typically French-derived) name (e.g. ``München'' rather than the English ``Munich'').

We then construct example sentences in English where each of our selected place names is used in four different contexts, corresponding to each of the four grammatical cases in Icelandic. (The exceptions are ``Paris'' and ``Berlin'', which are only tested in the genitive as they are practically the same as in English in the other three cases.) Our motivation is that due to the richer morphology of Icelandic, an accurate translation model needs to be able to map the same lexical form in English to several different forms in Icelandic, depending on the context (and this particular mapping is perhaps a problem better suited to older models than state-of-the-art LLM-based ones). 

We try to avoid the possibility that our sentences will be translated into Icelandic in a way that is generally correct but uses a different syntactic structure or wording than we anticipate, which would lend itself to the use of a different grammatical case than the one we intend to test for. We do this by keeping our example sentences short and simple and choose case-governing words and prepositions carefully to maximize the probability of a particular translation in Icelandic (e.g. the sentence ``The flight from Tórshavn to Gothenburg was delayed until the morning'' should almost certainly be translated using the prepositions ``frá'' and ``til'' for ``from'' and ``to'', governing the dative and genitive cases respectively.)

Given names, both in Icelandic and English, constitute the final part of our test suite. As in the case of the place names, we construct simple sentences in English containing Icelandic names and meant to test for each of the four grammatical cases. For this task, we chose a specific subset of common names in Icelandic: male-female pairings that take the weak inflection, e.g. ``Helgi''-``Helga'' and ``Gunni''-``Gunna'', where the male name has the ending -``i'' in the nominative case but -``a'' in oblique cases and the female name has the ending -``a'' in the nominative but -``u'' in the oblique cases (and possibly also a u-umlaut as in ``Svala'' $\rightarrow$ ``Svölu'').

These name pairs, of which we select 45 from the Database of Icelandic Morphology \cite{bjarnadottir-etal-2019-dim},\footnote{\url{https://bin.arnastofnun.is/DMII/}} are chosen as they seem to present a particular challenge for translation systems compared to names that take the strong declension. In constructing our test suite, we found that available models seemed to perform at random when asked to translate sentences containing these names in different cases, presumably due to the ambiguity of the lexical forms ending in -``a'', which can be a male name in an oblique case or a female name in the nominative. As oblique case nominals are a distinct and common feature of the Icelandic language \citep{hoski2007}, this problem is highly relevant in terms of correctly relaying the meaning of the sentence (see Table \ref{propernames-table}). 

\section{Results}
\begin{table*}
\centering
\begin{tabular}{lccccc}
\toprule
System & Total Idioms & Total Literals & Idiom Accuracy & Literal Accuracy & Total Accuracy \\
\midrule
AMI             & 100 & 65  & 0.29      & 0.892308   & 0.527273  \\
Aya23           & 93  & 49  & 0.0537634 & 0.122449   & 0.0774648 \\
Claude-3.5      & 96  & 56  & \textbf{0.75}      & 0.857143   & \textbf{0.789474}  \\
CommandR-plus   & 93  & 47  & 0.0967742 & 0.382979   & 0.192857  \\
CycleL          & 92  & 49  & 0         & 0.102041   & 0.035461  \\
Dubformer       & 91  & 53  & 0.340659  & 0.603774   & 0.4375    \\
GPT-4           & 93  & 48  & 0.430108  & 0.833333   & 0.567376  \\
IKUN-C          & 95  & 52  & 0.494737  & 0.75       & 0.585034  \\
IKUN            & 95  & 51  & 0.526316  & 0.607843   & 0.554795  \\
IOL\_Research   & 92  & 47  & 0.434783  & 0.702128   & 0.52518   \\
Llama3-70B      & 93  & 50  & 0.268817  & 0.62       & 0.391608  \\
ONLINE-A        & 188 & 107 & 0.265957  & 0.859813   & 0.481356  \\
ONLINE-B        & 102 & 69  & 0.22549   & 0.898551   & 0.497076  \\
ONLINE-G        & 97  & 66  & 0.185567  & 0.80303    & 0.435583  \\
TranssionMT     & 76  & 50  & 0.223684  & \textbf{0.92}       & 0.5       \\
TSU-HITs        & 92  & 48  & 0.0434783 & 0.104167   & 0.0642857 \\
Unbabel-Tower70B& 95  & 57  & 0.631579  & 0.877193   & 0.723684  \\
\bottomrule
\end{tabular}
\caption{\label{tab:idiom-manual}Results of manual evaluation of system performance on our idioms test suite. We randomly split up the translations of the test suite into segments of around 100 `idiomatic' example translations and around 50 `literal' example translations (see `Total' columns). The highest scores in each column are in bold. The authors reviewed the translations themselves and the reviewed examples, along with our grading, can be found at \url{https://github.com/stofnun-arna-magnussonar/idioms_names_test_suite/idioms/human_evaluation}.}
\end{table*}

\begin{table*}
\centering
\begin{tabular}{lccccc}
\toprule
System name & Total score & Correct idiomatics & CI ratio & Correct literals & CL ratio \\
\midrule
AMI & 0.447236& 83 & 0.21 & 184 & \textbf{0.9} \\
Aya23 & 0.169179& 39 & 0.1 & 62 & 0.3 \\
Claude-3.5 & \textbf{0.654941}& 216 & \textbf{0.55} & 175 & 0.86 \\
CommandR-plus & 0.293132& 66 & 0.17 & 109 & 0.53 \\
CycleL & 0.108878& 22 & 0.06 & 43 & 0.21 \\
Dubformer & 0.427136& 112 & 0.28 & 143 & 0.7 \\
GPT-4 & 0.547739& 161 & 0.41 & 166 & 0.81 \\
IKUN-C & 0.480737& 141 & 0.36 & 146 & 0.72 \\
IKUN & 0.509213& 161 & 0.41 & 143 & 0.7 \\
IOL\_Research & 0.482412& 133 & 0.34 & 155 & 0.76 \\
Llama3-70B & 0.417085& 99 & 0.25 & 150 & 0.74 \\
ONLINE-A & 0.442211& 86 & 0.22 & 178 & 0.87 \\
ONLINE-B & 0.447236& 85 & 0.22 & 182 & 0.89 \\
ONLINE-G & 0.413735& 71 & 0.18 & 176 & 0.86 \\
TranssionMT & 0.448911& 86 & 0.22 & 182 & 0.89 \\
TSU-HITs & 0.112228& 24 & 0.06 & 43 & 0.21 \\
Unbabel-Tower70B & 0.60804& 195 & 0.5 & 168 & 0.82 \\
\bottomrule
\end{tabular}
\caption{Results of automatic evaluation of system performance on our idioms test suite. We show the overall score for each system but also consider separately the percentage of idiomatic text examples marked as correct and the percentage of literals marked correct, to try and give an overview of the relationship between the two. Highest scores in each column are in bold. Our scripts for running automatic evaluation can be found at \url{https://github.com/stofnun-arna-magnussonar/idioms_names_test_suite/idioms}.}
\label{tab:idiom-auto}
\end{table*}

\begin{table*}
\centering
\begin{tabular}{lccc}
\toprule
System name & Total score & Total city score & Total people score \\
\midrule
AMI & \textbf{0.5399} & \textbf{0.4705} & 0.5861\\
Aya23 & 0.3838 & 0.0432 & 0.6103\\
Claude-3.5 & 0.5091 & 0.4591 & 0.5423\\
CommandR-plus & 0.3339 & 0.1205 & 0.4758\\
CycleL & 0.0 & 0.0 & 0.0\\
Dubformer & 0.4383 & 0.3614 & 0.4894\\
GPT-4 & 0.5109 & 0.2773 & \textbf{0.6662}\\
IKUN-C & 0.4691 & 0.2727 & 0.5997\\
IKUN & 0.4846 & 0.2886 & 0.6148\\
IOL\_Research & 0.4773 & 0.2205 & 0.648\\
Llama3-70B & 0.4138 & 0.3227 & 0.4743\\
ONLINE-A & 0.5345 & 0.4659 & 0.5801\\
ONLINE-B & 0.5109 & 0.4273 & 0.5665\\
ONLINE-G & 0.4065 & 0.3614 & 0.4366\\
TranssionMT & 0.5082 & 0.4227 & 0.565\\
TSU-HITs & 0.147 & 0.0932 & 0.1828\\
Unbabel-Tower70B & 0.5254 & 0.4114 & 0.6012\\
\bottomrule
\end{tabular}
\caption{Results of automatic evaluation of system performance on our names test suite, given as a proportion of properly scored city names `Total city score', properly scored given names `Total people score' and overall `Total score'. Highest scores in each column are in bold. Our scripts for running automatic evaluation can be found at \url{https://github.com/stofnun-arna-magnussonar/idioms_names_test_suite/names}. (Note that the zeroes for CycleL's submission are not a mistake, this submission performed poorly and our scoring strategy is not particularly forgiving.)}
\label{tab:name-auto}
\end{table*}

All submissions were scored using automatic metrics we constructed. Furthermore, we manually reviewed around 150 randomly selected examples in the case of the idioms (around 100 `idiomatic' examples and around 50 `literal' examples for each submitted system). The authors reviewed the translations themselves, manually changing the scores given by our automatic method (using the `positive' and `negative' keywords discussed in \ref{idmethod}) if they deemed it wrong. 

The translations of our proper names suite was only carried out with naive automatic methods. The translations were lemmatized using a lemmatizer for Icelandic \cite{ingolfsdottir-etal-2019-nefnir} and compared with a reference of which Icelandic lemmas should appear in the translation and in which grammatical form (being able to look up lemmas is especially useful for the given names, since the male and female names share surface forms).

We show the results of our manual evaluation in Table \ref{tab:idiom-manual} and the results of automatic metrics for our idioms test suite in Table \ref{tab:idiom-auto}. For our names test suite, we show the results of our automatic metrics in Table \ref{tab:name-auto}. Our scripts for running the automatic evaluations and the manually reviewed examples are released along with our test sets.

\subsection{Scores for Idiomatic Expressions}

Our results show a wide range of performance across different models. The best overall accuracy on the idioms test suite is achieved by Claude 3.5, with Unbabel-Tower70B a close second, as indicated both by our automatic and manual evaluation. Claude 3.5 is also the highest-scoring submission when we only consider translations of expressions used in an idiomatic sense, both according to our automatic metrics and the manual review, and Unbabel-Tower70B the clear runner-up.

When considering the literal translations in isolation, however, the overall two best models are narrowly `beaten' by a few models that score considerably lower overall. According to our automatic metrics, our own submission (AMI) scores highest in that category, only slightly ahead of ONLINE-B and TranssionMT. These three also come out on top in the manual evaluation, with TranssionMT recording the highest score (a superb 0.92) and ONLINE-B and AMI following in second and third. 

This discrepancy between performance in translating phrases in an idiomatic context and a literal context is very interesting - these three models all scored under 0.3 in idiomatic accuracy, which suggests that for some models, proficiency in effectively translating text in a literal sense comes at a cost to their ability to handle more metaphorical text. The best-performing models overall, however, were seemingly able to maneuver quite effectively between both use cases. Models, perhaps predictably, generally score higher when translating literal usage than when translating idioms.

\subsection{Scores for Proper Names}

In terms of the proper names suite, place names prove to be much more difficult for the submitted models than people's names. It is the submission by our own team which narrowly tops the list overall, ahead of ONLINE-A and Unbabel. The AMI submission also ranks highest when place names are considered in isolation, although it still gets fewer than half of all names correct. For given names, GPT-4 scores highest.

For this part of our test set, we report no manual evaluation. A cursory glance at the output, however, shows that our naive automatic scoring method still leaves quite a bit to be desired. A problem with testing for specific grammatical forms in each case is that the correct form can change depending on the sentence structure. As discussed in \ref{propernames}, we tried to control for this by keeping test sentences brief and unambiguous. Even so, we find there are examples of different phrasings than we expected in some translation outputs that call for a different grammatical form of a name than our scoring mechanism supposes, but can still be considered a decent translation. 

This especially applies to the sentence form: X ``cares for'' Y. We assumed a correct translation into Icelandic would be: X ``þykir vænt um'' Y, where X would take the dative case and Y the accusative. The submitted systems, however, had many different ideas on how best to phrase this system, not all of them completely wrong. 

We therefore recognize that our scoring system needs to be fine-tuned but nevertheless believe the very low scores are mainly a reflection of the difficulty of this task.

\section{Conclusions and Future Work}

Scores on both sets are relatively low, indicating that these particular categories continue to pose some problems for even state-of-the-art translation models and that there is considerable room for improvement. 

Future work can explore further comparison of performance and fine-tuning of our automatic scoring methods. Given time, we could also have investigated whether more manual evaluation, ideally using more evaluators, would have resulted in different scores.

We also note that our test suite can be adapted with relative ease into other languages and hope that this allows for further work on other language directions.

\section*{Limitations}

There are several judgment calls to be made when working with our chosen categories and many of the decisions we made in terms of selecting items to be translated, defining automatic metrics for `right' and `wrong' translations and manual evaluation can be argued for or against. We are aware that the choices we make could be indicative of potential biases of the authors and that a different team, perhaps with a different demographic makeup, might well have constructed the test set and evaluated the translations in a different way.

These necessary choices are perhaps most apparent in terms of our idioms set. Evaluation of linguistic acceptability of translations and correspondence of idiomatic phrases between languages is based on our intuition and we are aware that fluent speakers of English and Icelandic may disagree on some decisions. Another point to consider is the degree to which we want our test set to be prescriptive - as a simple search on the Internet can prove, there are multiple usages of common English idioms directly translated into Icelandic, e.g. on social media \citep{Hilmisdóttir_Huhtamäki_Karlsson_2023}. Determining at what point to say this usage is no longer `incorrect' is an interesting question of ethics and philosophy of language.

As for our set of proper names, there exists some speaker variation in how and when place names are translated into Icelandic, although we have tried to limit our set to fairly uncontroversial choices (see discussion in \ref{propernames}). The requirement of not translating English names into Icelandic is less cut and dried, as it may be appropriate for a machine translation model in some cases, e.g. in literary text or the discussion of royal or historical figures. It can also be noted that some of our English names are, in fact, given names in Iceland. This should not affect our results, however, as we allow for the inflection of a final -``a'' into -``u'' in female names like ``Pamela'' and in other cases, `non-Icelandic' names typically remain completely unchanged in all grammatical cases.

\section*{Acknowledgments}

We would like to thank the WMT24 test suite organizers for the assistance and communication while working on this submission and two anonymous reviewers for helpful feedback.

\bibliography{emnlp2023}
\bibliographystyle{acl_natbib}

\end{document}